# Deep Learning-Based Inverse Design for Engineering Systems: Multidisciplinary Design Optimization of Automotive Brakes

Seongsin Kim[1]

Department of Mechanical Systems Engineering,

Sookmyung Women's University, Seoul, Korea

kss@sm.ac.kr

Minyoung Jwa[1]

Department of Mechanical Systems Engineering,

Sookmyung Women's University, Seoul, Korea

minyoung306@sm.ac.kr

Soonwook Lee

Hyundai Mobis, Seoul, Korea

lsw@mobis.co.kr

Sunghoon Park

VEng, Seoul, Korea

shpark@veng.co.kr

Namwoo Kang*

The Cho Chun Shik Graduate School of Green Transportation,

Korea Advanced Institute of Science and Technology, Daejeon, Korea

nwkang@kaist.ac.kr

[1]Contributed equally to this work.

*Corresponding author

# Abstract

The braking performance of the brake system is a target performance that must be considered for vehicle development. Apparent piston travel (APT) and drag torque are the most representative factors for evaluating braking performance. In particular, as the two performance factors have a conflicting relationship with each other, a multidisciplinary design optimization (MDO) approach is required for brake design. However, the computational cost of MDO increases as the number of disciplines increases. Recent studies on inverse design that use deep learning (DL) have established the possibility of instantly generating an optimal design that can satisfy the target performance without implementing an iterative optimization process. This study proposes a DL-based multidisciplinary inverse design (MID) that simultaneously satisfies multiple targets, such as the APT and drag torque of the brake system. Results show that the proposed inverse design can find the optimal design more efficiently compared with the conventional optimization methods, such as backpropagation and sequential quadratic programming. The MID achieved a similar performance to the single-disciplinary inverse design in terms of accuracy and computational cost. A novel design was derived on the basis of results, and the same performance was satisfied as that of the existing design.

**Key Words:** Deep Learning, Multidisciplinary Design Optimization, Inverse Design, Brake System

# 1. Introduction

Among the many factors for evaluating a vehicle's performance, braking force, which is directly related to safety, is considered to be extremely important. In addition, braking force not only determines the safety performance of a vehicle but also affects its fuel efficiency, ride behavior and handling. Therefore, designating a brake system with a good braking force is an essential process in vehicle design. The brake system that determines the braking force is typically classified into disc brake, drum brake, and engine brake, in which the classification depends on the operating principle and the position of action. Moreover, the disc brake system consists of a principle in which braking force is generated when the friction caused by the brake pad pushes the disc rotor, and then this force combines with the friction of the wheel and rotated by hydraulic force (Kinkaid et al., 2003). For this disc brake system, the two main detailed performances affecting the braking force are apparent piston travel (APT) and drag torque.

APT is the piston's traveled distance, and it is determined on the basis of the required amount of braking oil while the brake is applied (Antanaitis & D. B., 2013). Drag torque is a generated unnecessary braking torque by the piston when the braking system presses the brake pads without fully returning to the existing position (Reich et al., 2015). These performances have a tradeoff relation to the piston's roll-back distance. The longer the roll-back the piston has, the longer the piston travels at the next braking, suggesting a decrease in APT performance. However, after the brake is released, the force in which the piston pushes the brake pads is reduced. In such cases, the generated residual torque is reduced, and the drag torque performance is increased. Although a brake system for satisfying both APT and drag torque performances is difficult to design, the brake system must still adequately satisfy both aspects in a simultaneous manner (Liles et al., 1989).

Researchers have used various design variables of the disc brake system, including the caliper, rotor, brake pad, seal groove, etc. Among them, the seal groove's design determines the movement of the seal during braking, which greatly affects the piston's movement distance. Cai et al. (2002) recommend the seal groove design as the cheapest approach of adjusting the performance of the brake system, and much progress has been achieved in the performance analysis of seal groove designs ever since. This study contributes to this research domain by proposing an optimized design of the seal groove to simultaneously satisfy the desired APT and drag performances.

A multidisciplinary design optimization (MDO) approach is required for the brake design to achieve this study's purpose. However, the computational cost of MDO increases as the number of disciplines increases. Recent deep learning (DL)-based inverse design studies have shown the possibility of rapidly creating an optimal design to satisfy target performance without implementing an iterative optimization process (Liu et al., 2018; Nomura et al., 2019; Luo et al., 2020). The inverse design is a target-oriented design methodology of designating the desired target parameters and satisfies them; it is a design methodology of inversely discovering an optimized design that does not exist in an established database (Molesky et al., 2018). The existing DL-based inverse design studies focus on nanophotonics and micromaterials (Peurifoy et al., 2018; Pilozzi et al., 2018; Zhang et al., 2019). The research on engineering system design at the system level is generally insufficient. Our research centers on a new design optimization methodology with high accuracy and low computational cost, and it is developed by integrating the DL-based inverse design methodology and MDO for the engineering design research field. To the best of our knowledge, our study is the first DL-based inverse design research for addressing the MDO problems of automotive systems.

This work proposes a multidisciplinary inverse design (MID) framework to achieve the target APT and drag performances for brake systems. The MID framework consists of three stages. The first stage is the data analysis and preprocessing process. This stage includes seal groove design parameterization and simulation. The second stage is the exploration of DL architectures for APT and drag performance prediction. This process involves the building of a surrogate model for design optimization for the third stage. The third stage is the building of an MID model, which can generate the optimal design for satisfying the target APT and drag performances.

The rest of this paper is organized as follows. Section 2 reviews the previous works. Section 3 introduces the proposed MID framework, and Section 4 describes a preprocessing process. In Section 5, DL models for performance prediction are introduced. Section 6 and 7 describe the single-disciplinary inverse design (SID) model for APT performance and the MID model for both APT and drag performances, respectively. Then, the comparative results of the DL-based inverse design and the traditional optimization methods are analyzed. Section 8 summarizes the conclusion and future research plan.

## 2. Related Works

### 2.1 APT and Drag Torque in a Brake Design

Fig. 1(a) shows a typical disc brake system consisting of the caliper, rotor, and brake pads. When braking is applied, the rubber seal seals the bore of the piston and retracts the piston with storing energy (Anwana et al., 2002), as shown in Fig. 1(b). When the piston moves on the basis of the hydraulic force, it applies pressure to the brake pads, and a resistance of the seal is generated. With the release of the pedal, it pulls the piston back and releases energy. However, for braking forces exceeding a certain level (e.g., values higher than the resistance), the gap between the pad and the disc is not restored to its initial state. The less these parts are restored, the higher the generated drag torque. A larger piston retraction leads to a larger piston displacement, which then increases the braking time and limits the distance due to larger pedal travel (Phad et al., 2015). Thus, The APT values and drag torque amplification can be derived on the basis of the behavior of the rubber seal during braking (Anwana et al., 2002). Numerous finite element analysis (FEA) methods have been used to simulate the behavior of the seal groove of the piston against pressure (Ingale et al., 2016). Fig. 1(c) shows the behavior of a pressed seal groove via FEA simulation.

APT and drag performances affect the overall performance of vehicles. Piston travel distance determines the required time for braking. That is, the APT performance determines the reaction rate of the pedal, as "felt" by a driver when braking. When the APT performance value is high (i.e., the piston entails a long travel distance for braking), the required braking time is more extended, and the driver feels the "slope" of the pedal. The "feel" of the brake depends on APT performance (Ho et al., 2015). Many studies have proposed brake system designs to improve the braking feeling (Day et al., 2009; Celentano et al., 2004).

Drag torque, which means unnecessary residual torque, may shorten the lining's life, increase the fuel consumption, and lead to energy loss (Tao et al., 2003). Drag torque amplification should be reduced to improve fuel efficiency and lengthen the life of vehicles. Many simulation models have been developed to analyze brake systems with drag torque issues. Tao et al. (2003) developed a 1D simulation model covering all the significant factors and determined the impact of each parameter to the drag torque occurrence. Tamasho et al. (2000) presented a brake system design concept for reducing brake drag torque in the non-braking mode.

As APT and drag torque amplification are derived during piston retraction and relaxation, the stiffness,

material, and size of many parts of the brake system must be considered in the design process to satisfy the APT and drag performances. Many studies have also proposed effective friction materials and pad lining materials to improve brake performance (Doi et al., 2000; Aoki et al., 1980).

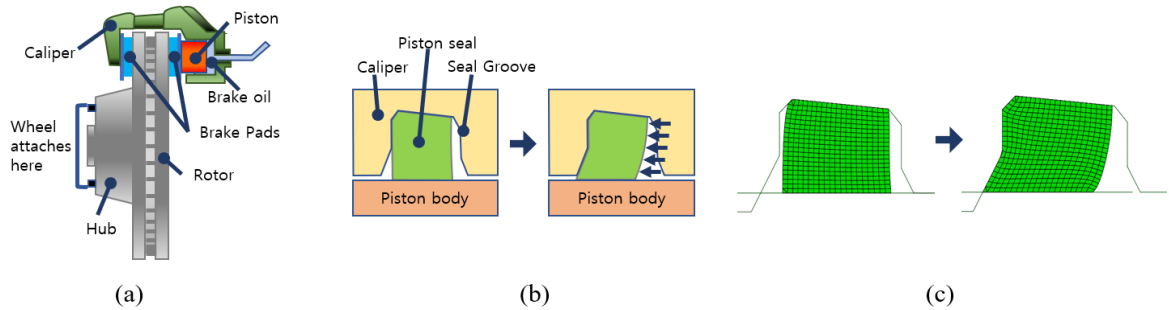

Fig. 1. Disc brake system structure and change in seal groove after hydraulic pressure: (a) structure of the disc brake assembly (Nathi et al., 2012); (b) behavior of the piston seal under pressure; and (c) behavior of the piston seal under pressure on FEA simulation.

2.2 Inverse Design via DL

The purpose of design optimization is to design a system with maximum performance while satisfying a given constraint. Traditional design optimization methods are largely divided into two fields. First, gradient-based methods, such as sequential quadratic programming (SQP) (Gill et al., 2005), perform optimization by using differential values for the design variables of the objective function and the constraints. Second, heuristic methods, such as genetic algorithms (Whitley & D., 1994), explore and exploit the design space without the use of differential values. However, the conventional methods that adopt the iterative optimization process entail high computational costs when finding designs in the high-dimensional design space.

As an alternative, DL has recently been used for design optimization. DL has the advantage of discovering complex patterns in large amounts of data. For design optimization, DL has been mainly used to build surrogate models (metamodels) (So et al., 2020; Messner et al., 2020). The DL studies for improving the design exploration performance of topology optimization-based generative designs have been widely conducted (Yang et al., 2018; Oh et al., 2018; Oh et al., 2019; Kallioras & Lagaros, 2020; Sun & Ma, 2020; Yoo et al., 2021). Yang et al. (2018) mapped latent variables and microstructures by using generative adversarial networks (GANs), and a microstructure with the desired properties was obtained with the application of the Bayesian optimization framework. Deep generative design (Oh et al., 2019), which is the basis of this study, generates novel designs similar to the actual designs by integrating the topology optimization and generative model of DL. Kallioras and Lagaros (2020) integrated reduced-order models and convolution filters of deep networks to generate various topology designs. Sun and Ma (2020) proposed a reinforcement learning-based generative design that would not require preoptimized topology data. Yoo (2021) proposed a methodology for generating 3D CAD designs and evaluated engineering performance by using a topology optimization framework.

In the engineering field, DL has mainly developed into a role of predicting performance by using design parameters. However, for actual problems, training models not only should be tested for output (performance) prediction in relation to a given input (design) but also instantly determine the optimal input (design) corresponding to a specific output (performance). DL-based inverse design is a design methodology of finding the optimal design (i.e., the design had not existed in the database) in real time while satisfying target performance. DL has been widely used in nanophotonics fields, such as synthesis paths of organic compounds, optimization of solar power generation, redox flow batteries, and a variety of other solid materials (Cova et al., 2019; Khazaal et al., 2020; Sanchez-Lengeling et al., 2018, Noh et al., 2020).

DL-based inverse design can be classified into supervised learning and unsupervised learning, in which the classification is based on whether label data are used. Supervised learning approximates the function of mapping an input to an output based on the input–output pair (Caruana et al., 2006). A design optimization problem can also have a pair of target values for the design value. However, no single design value can represent the target value because of the existence of various designs. For example, in the design problem of generating a groove seal shape, when the target performance of APT is inserted, not only one seal shape will correspond to one target value, but countless seal shapes will also appear; however, this is not an explicit one-to-one issue. We expect the values of the input to be mapped to those of the output, but learning will likely fail because of the absence of unique mapping. Here, we solve this problem by adopting the inverse design methodology of using the forward network trained to predict the target value of the design (Liu et al., 2018).

An example is shown in Fig. 2(a). We expect the value of $x$ to be mapped to $y$, but learning is not pursued because of the lack of a 1:1 correspondence. By attaching a forward network via a pretrained performance prediction model to the inverse design (Fig. 2(b)), $y$ can be entered as an input, pass through the forward network, and exit as an output. Training can reduce the difference between $y$ and target value $\hat{y}$ by freezing the forward network. Then, the optimal design $x$ can be extracted from the intermediate layer.

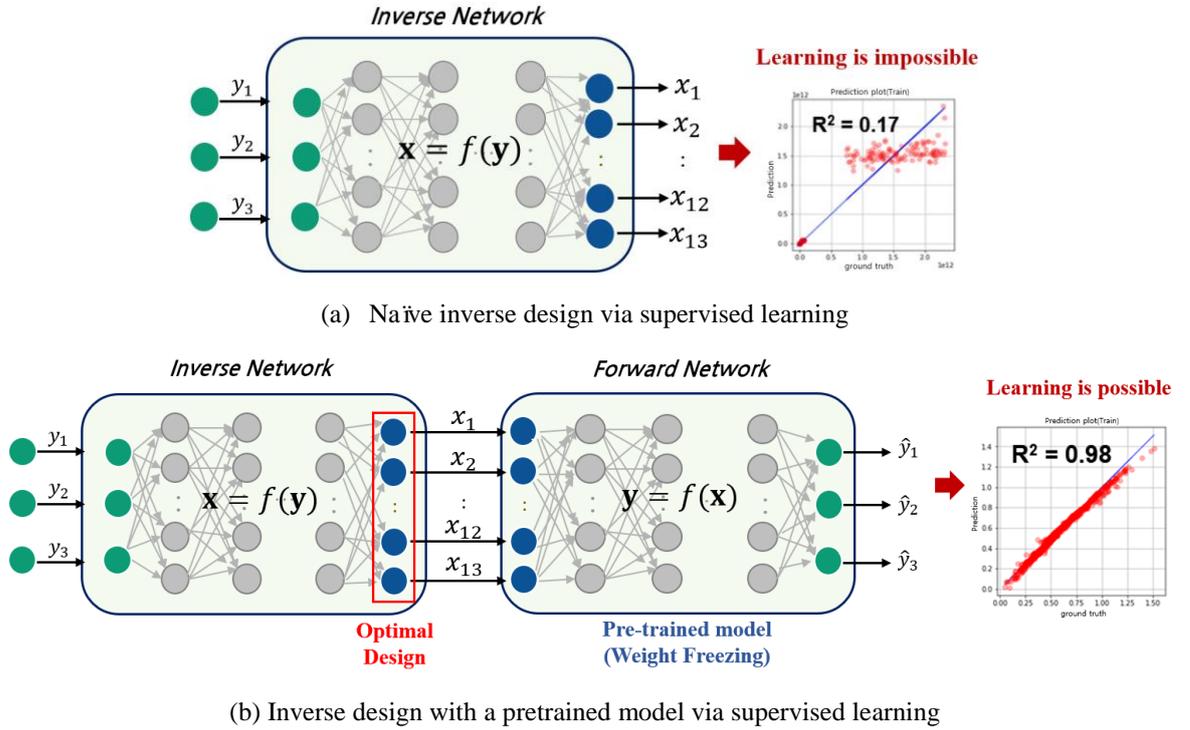

(a) Naïve inverse design via supervised learning

(b) Inverse design with a pretrained model via supervised learning

Fig. 2. Concept of inverse design via supervised learning.

Recent studies on the inverse design with supervised learning models include a comparison of known label values for specific optical properties (Peurifoy et al., 2018) and a complex electromagnetic scattering instance training dataset in which a deep neural network (DNN) was utilized. Some researchers (e.g., Liu et al. 2018) have provided a methodology for designing photonic structures. Pilozzi et al. (2018) introduced an inverse design methodology that uses an artificial neural network to obtain the complex parameters of an optical topological insulator at a desired target frequency.

In contrast to supervised learning, unsupervised learning processes data by ignoring explicit labels. The network infers important design patterns from the data without considering the desired answers or labels (Liu et al., 2018; Ma et al., 2019; Noh et al. (2019). The research on inverse design that uses unsupervised learning models include the generation of optical spectra of metasurfaces by using GANs or variational auto-encoders and the identification of materials with a set of desired properties. Research has also been conducted on efficiently generating metamaterials with respect to the target optical reactions and obtaining their spectra; since then, unsupervised learning has been mainly used to identify new structures and consequently address various engineering issues (Liu et al., 2018; Dan et al., 2020; Ma et al., 2019; Noh et al., 2019; Jiang et al., 2019).

Inverse design has been extensively used in the science field, such as nanophotonics and material design.

However, DL-based inverse design is rarely applied when solving system-level engineering design problems.

Sekar et al. (2019) used deep convolutional neural networks (CNNs) for an inverse design model in the domain of airfoil design. Different from the traditional regression model, the position where the design shape and performance could be derived was altered by swapping two performance parameters. For example, the coefficient of pressure (Cp) distribution of the airfoil was taken as the input to the CNN, whereas the airfoil shape was obtained as the output. The simple swapping of parameters is applicable in airfoil design, but this change is difficult to implement in all domains. This difficulty can be attributed to the non-application of the fundamental property of design (i.e., the performance matching must be unique).

Ghosh et al. (2022) demonstrated the ability of a probabilistic inverse design machine learning (PMI) methodology to optimize an industrial gas turbine. The methodology entailed a complex architecture in which different methodologies, such as the Gaussian process and a conditional invertible neural network, are combined. This architecture showed good performance but may be inapplicable to other domains.

Our study shows that inverse design can be applied in design optimization at the engineering system level and is particularly expandable to the MDO. The effectiveness of the methodology was verified using an actual brake system design.

## 3. Proposed Framework

The proposed framework of this study is shown in Fig. 3. The framework is divided into three stages. Data preprocessing is performed at the first stage. On the basis of the prepared data, performance prediction models are developed in the second stage. Finally, inverse design modeling is performed on the models in stage 3.

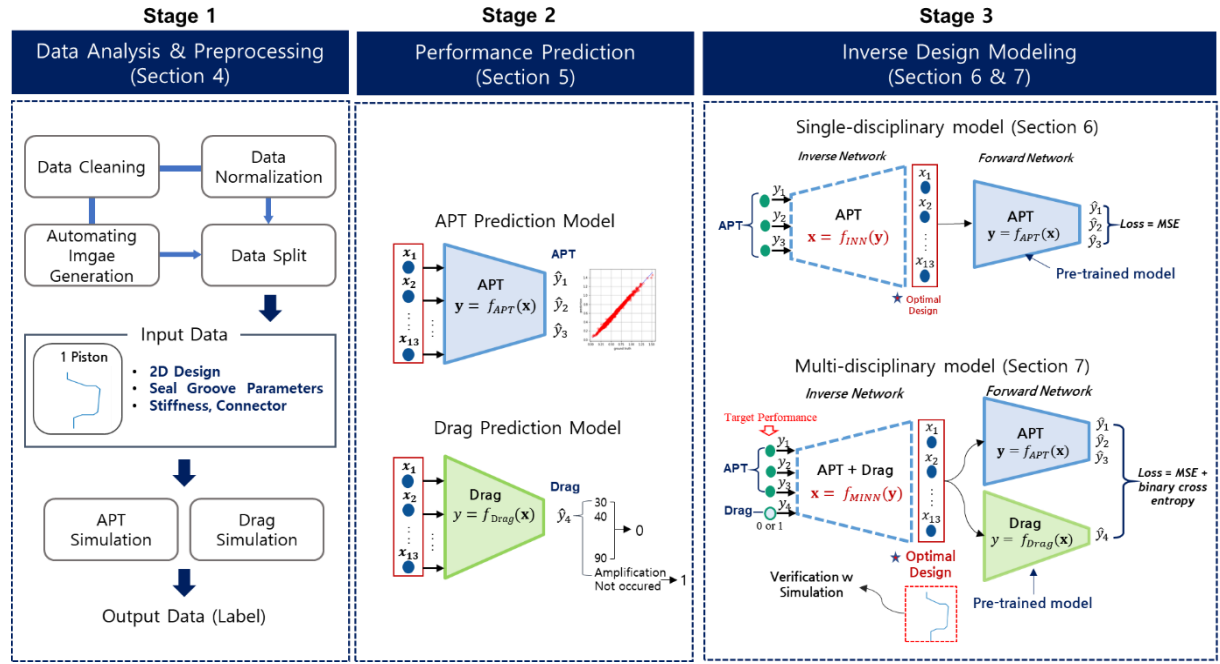

Fig. 3. DL-based inverse design process of brake systems.

● Stage 1: **Data preprocessing** – The piston sea groove geometry is parameterized, and design variables are defined. Design of experiment (DOE) and engineering simulation are conducted to collect the labeled data. For the training performance prediction model, we processed data cleaning and normalization.

● Stage 2: **Performance prediction** – A surrogate model is developed to predict the APT and drag performances. The DNN and CNN models are trained, and the optimal architecture and hyperparameters are explored.

● Stage 3: **Inverse design modeling** – The SID models and the MID model are developed on the basis of the pretrained surrogate models from stage 2. Finally, the optimal design with the given target performance is generated in real time.

Sections 4 and 5 describe stages 1 and 2, respectively. Sections 6 and 7 introduce the SID and MID for stage 3, respectively.

## 4. Data Analysis and Preprocessing (Stage 1)

### 4.1. Design Parameterization

For the APT and drag performance interpretations, 25 design factors were used as design variables. The variables were segmented into the following groups: seal shape design group (20), calipers and pads' stiffness group (2), and setting guide pin bush movement group (3). We extracted 13 key input variables to predict the APT and drag torque performances. Then, the extracted data were subdivided into four groups: seal shape group (7), piston and seal's size group (2), stiffness group (3), and connector group (1). The details are shown in Table 1.

Table 1. Thirteen key input variables divided into four groups.

| Group name | Name of design variables |
| --- | --- |
| Seal groove | chamfer ($x_1$), param_y_angle ($x_2$), param_z ($x_3$), param_a ($x_4$), param_d ($x_5$), param_e ($x_6$), param_p_r ($x_7$) |
| Piston and seal size | seal_height ($x_8$), param_s_t ($x_9$) |
| Stiffness | caliper_stiffness ($x_{10}$), pad_stiffness ($x_{11}$), gp_bush_stiffness ($x_{12}$) |
| Connector | gp_bush_load_limit($x_{13}$) |

The seven variables of the "seal" group are shown in Fig. 4. On the basis of the seven variables related to the seal shape, 14 points were calculated. These 14 points represent the vertices and imaginary points depicting a seal groove geometry. As each point is properly connected, the seal groove geometry can be represented, as shown in Fig. 4.

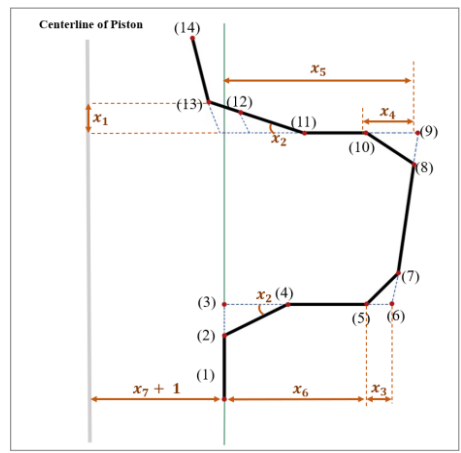

Fig. 4. Seal groove geometry with 14 points calculated from 7 variables $(x_1, x_2, \ldots, x_7)$.

The process of deriving the APT and drag performances was implemented via ABAQUS simulation analysis (Giner et al., 2009). Thirteen variables were analyzed (Sections 4.2 and 4.3).

## 4.2 APT Simulation

Here, APT performance is defined as the travel distances of the piston when three different pressures are applied (Anwana et al., 2003). The initial position and the positions at different pressures were considered when calculating the travel distance. The APT performance in terms of low, medium, or high pressure of the piston was analyzed using ABAQUS. The process can be described as follows.

First, in setting up the analysis environment, the seal groove model was assembled on ABAQUS Then, we released the stress of the brake and stabilized the assembled seal. In the process of stabilizing the seal, we preloaded the very low braking pressure with $n$ times. Then, the brake system was operated with high pressure. (See Table 2) The APT performance was analyzed by extracting the piston displacement values for the three different pressures (4%, 40% and 100% of the maximum pressure representing low, medium, and high). As shown in Fig. 5, piston displacement values ($r_{init}, r_{4\%}, r_{40\%}, r_{100\%}$) are extracted in the braking process. $r_{4\%}, r_{40\%}$ and $r_{100\%}$ are the positions of the piston when 4%, 40% and 100% of maximum hydraulic pressures are applied. Also, $r_{init}$ is the piston's initial state position without pressure. APT performances ($APT_1, APT_2, APT_3$) are defined as the difference between the position of the piston at low, medium and high pressure ($r_{4\%}, r_{40\%}, r_{100\%}$) and the initial position of the piston ($r_{init}$). That is, the three APT performances represent the piston's traveled distance under different pressures.

Table 2. APT performances extracting process.

| Pressure | Operating process |
| --- | --- |
| Very low pressure with $n$ times | No APT values are measured (pre-setting) |
| High pressure | APT values are measured for 4%, 40% and 100% pressures. |

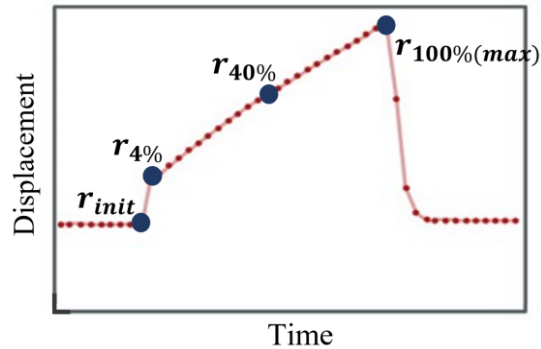

Fig. 5. Interpretation graph for extracting APT performance.

## 4.3 Drag Simulation

In general, the higher the braking pressure, the higher the drag torque and the better the drag performance, with the drag torque slowly increasing. The ideal drag performance is to maintain the drag torque to a constant value. The drag torque for the braking pressure may be adjusted according to the design of the brake system. In this study, the roll-back distance of the gap between the disc and pad is used as an indicator to replace drag performance. The smaller the roll-back distance (the larger the gap), the higher the drag torque for a brake system. Therefore, the drag performance can be determined depending on how high the inflection point is, at which the roll-back distance can rapidly decrease (the drag torque rapidly increases).

Fig. 6 shows an ABAQUS-based interpretation graph. The pressure gradually increases over time and outputs a roll-back distance when braking is applied and released. The process of analyzing in ABAQUS and deriving the drag performance value can be described as follows.

First, we processed the interpretation for drag torque in ABAQUS by gradually increasing the pressure from 10 to 100 bar for the disc brake system model. We applied a wide range of pressure to obtain the spectrum. The positions of the disc and brake pads were determined as we applied braking and after braking. The roll-back distance was extracted on the basis of the difference in positions of the disc and brake pad. The point at which the roll-back distance decreases rapidly was defined as the point at which the roll-back distance is lowered by more than 5% from the average roll-back distance of the initial pressure (i.e., 10, 20, and 30 bar). In the case of Fig. 6, the braking pressure reduced by 5% with respect to the average roll-back distance at 10 to 30 bar is "60" bar. As the objective was to extract the inflection of roll-back distance, we defined the previous pressure (i.e., "50" bar) as the drag performance.

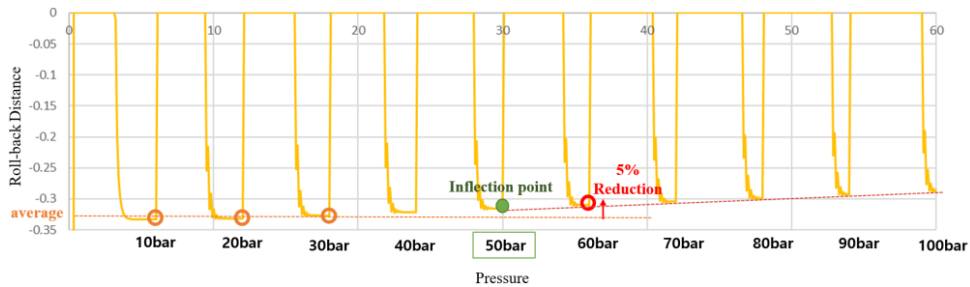

Fig. 6. Interpretation graph for extracting drag performance.

### 4.4 DOEs

In this work, for the DL model's training, we generated the input data for each APT and drag performance via Latin hypercube sampling techniques and extracted the output data via interpretation experiments in ABAQUS (Sections 4.2 and 4.3). In the process of extracting the output, the final number of data was decreased due to failure of interpretation.

The total data of APT and drag performance were 820 and 1691, respectively. The distribution of the total dataset is shown in Fig. 7. Fig. 7(a) shows the distribution of the APT dataset, and Fig. 7(b) shows the distribution of the drag dataset.

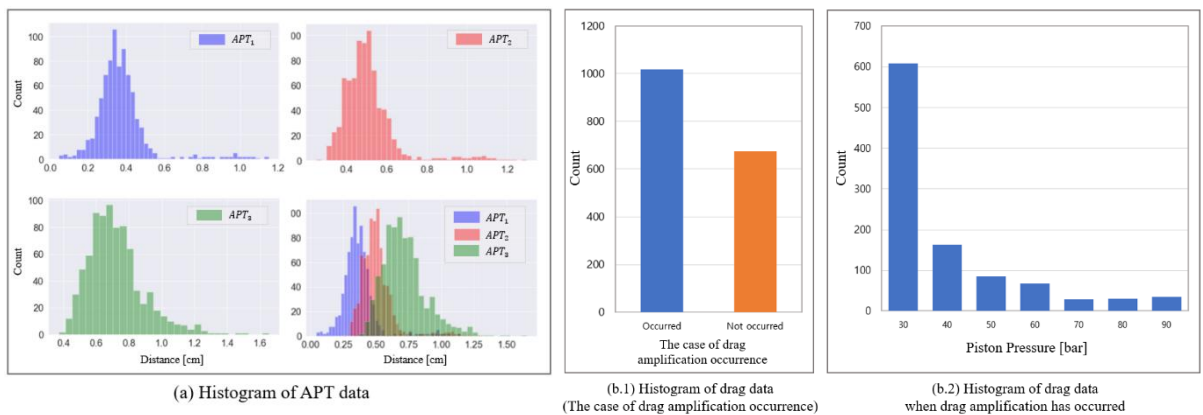

Fig. 7. Histograms of the brake performance datasets.

4.5 Data Preprocessing

The process of preprocessing the generated data into a form capable of performing DL can be described as follows.

First, during the ABAQUS analysis experiment, data cleaning was performed to delete the missing data caused by convergence failure. The model training was smoothed via "Min-Max Scaling" normalization, a tool for adjusting the data to a value between 0 and 1. Second, we developed an automation code for creating seal groove images based on the design factors for CNN model learning in MATLAB. On the basis of the design values, 14 points constituting the seal groove could be calculated. Using the 14 points, the seal groove's cross-sectional images were generated and preprocessed as binary images. The shape of the individually generated images was 204 × 204. A code was developed to automate the image generation, subsequently obtaining approximately 2500 images. Finally, a 7:2:1 ratio was selected to classify the data for model training, verification, and testing. The final training data were labeled as Train Data (574 types), Validation Data (164 types), and Test Data (82 types). The final training data types are summarized in Table 3.

Table 3. Dataset size of APT and drag performances.

| Property | Training | Validation | Testing |
|---|---|---|---|
| APT | 574 | 164 | 82 |
| Drag | 1183 | 339 | 169 |

# 5. Performance Prediction (Stage 2)

In stage 2, APT and drag performance prediction models are developed using preprocessed data. In this study, DNN and CNN models were trained to identify the optimal architecture of each performance variable.

5.1. APT Prediction

The input data varied depending on the architecture of the DNN and CNN models. In the case of the DNN model, only 13 design parameters were used as the input, whereas an additional 2D seal groove images preprocessed for learning features were used in the CNN model. Both models were developed in the direction of minimizing the loss. The mean-squared error (MSE) between the actual and predicted APT performances si calculated as

$$MSE = \frac{1}{n}\sum_{i=1}^{n}(\hat{y}_i - y_i)^2. \qquad (1)$$

A hyperparameter search was performed to aid in the development of a high-performance prediction model. Figs. 8(a) and 8(b) are visualizations of the DNN and CNN models with DL architectures, respectively. The learning rate of DNN was set to 0.0005, and the batch size was set to 128. The Adam optimizer was used, and an "early stop" function was applied to prevent overfitting. The CNN model in this study comprised five convolutional layers, four max pooling layers, a ReLU activation function, and latent space layers. The newly added regressor part consisted of seven fully connected layers. Fig. 9 shows the training curves of the two models. For the CNN model, the training time was approximately 15 minutes on four GTX 1080 parallel GPUs; for the DNN model, it was approximately 5 minutes.

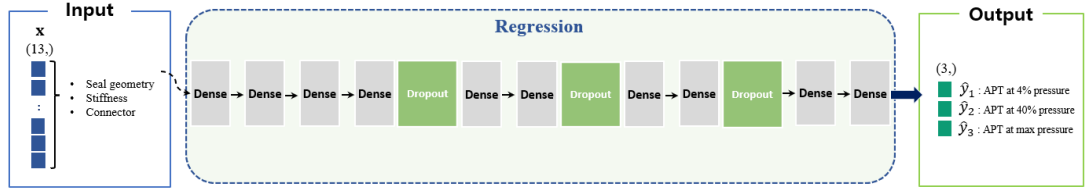

(a) Architecture of DNN Model

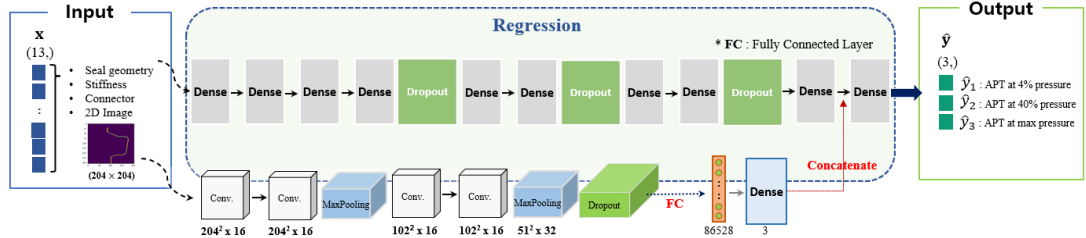

(b) Architecture of CNN Model

Fig. 8. Architecture of the APT performance regression model.

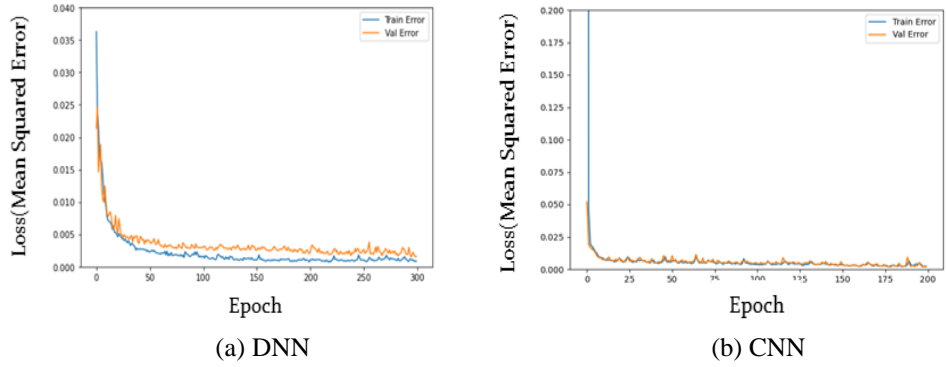

(a) DNN                    (b) CNN

Fig. 9. Training curve (loss) of the DNN and CNN models for APT.

The three evaluation metrics used to evaluate the performance of the predictive model were the root mean square error (RMSE), mean absolute error (MAE), and R square ($R^2$). RMSE and MAE are defined as follows:

$$RMSE = \sqrt{\frac{1}{n}\sum_{i=1}^{n}(\hat{y}_i - y_i)^2} \quad , \tag{2}$$

$$MAE = \frac{1}{n}\sum_{i=1}^{n}|\hat{y}_i - y_i| \quad , \tag{3}$$

where $\hat{y}$ is the predicted value, y is the ground truth value, and n is the data size. Table 4 shows the results of testing the two models. Our baseline model was a fully connected DNN model with one hidden layer. The hidden layer comprised 128 neurons. The performance of the DNN model was better than that of the CNN model. Thus, the DNN model was selected as the final model.

Table 4. Comparative prediction results of APT performance

| Methodology | Test | | |
| --- | --- | --- | --- |
| | MAE | RMSE | $R^2$ |
| Baseline | 0.08 | 0.10 | 0.79 |
| DNN      | 0.04 | 0.06 | 0.90 |
| CNN      | 0.05 | 0.06 | 0.85 |

## 5.2. Drag Prediction

The drag classification model was built on the DNN network, and the total training dataset was paired with 1691 data and labels. The drag data was divided into cases as to whether drag torque occurs or does not occur. As the cases needed to be considered separately, we developed a two-stage drag prediction model.

The first model (first classification model) entailed a binary classification, i.e., whether drag amplification occurs or does not occur. Meanwhile, the secondary model (second classification model) was used to classify the values of the seven classes (e.g., 30 to 90) when only amplification occurs. In the secondary model, one-hot encoding for the data to define each class. Fig. 10 shows the conceptualization of the first and second classification models.

Figs. 11(a) and 11(b) are visualizations of the DNN model architecture of each model. As shown in Fig. 11(b), the one-hot encoded data (i.e., drag amplification occurs) have seven dimensions. The "argmax" function was used to derive the final single prediction class. The loss function was defined as the binary cross entropy for the first model and the cross entropy for the second model. We used an Adam optimizer with corresponding learning rates of 0.0055 and 0.0002. For the performance evaluation of the predictive model, accuracy (%) was used as the evaluation metric. Table 5 shows the result of the fine-tuned model. For the design optimization, only the first classification model was used to determine a design that is free of drag.

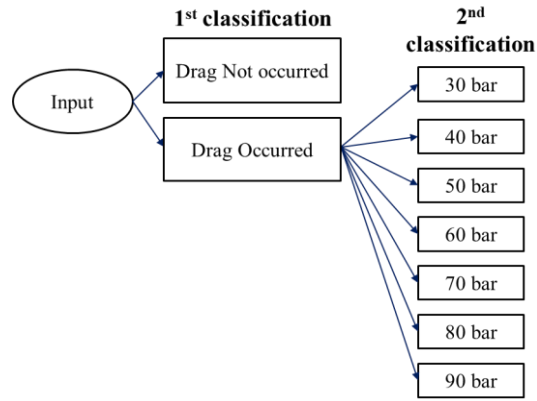

Fig. 10. Conceptualization of drag performance for model classification.

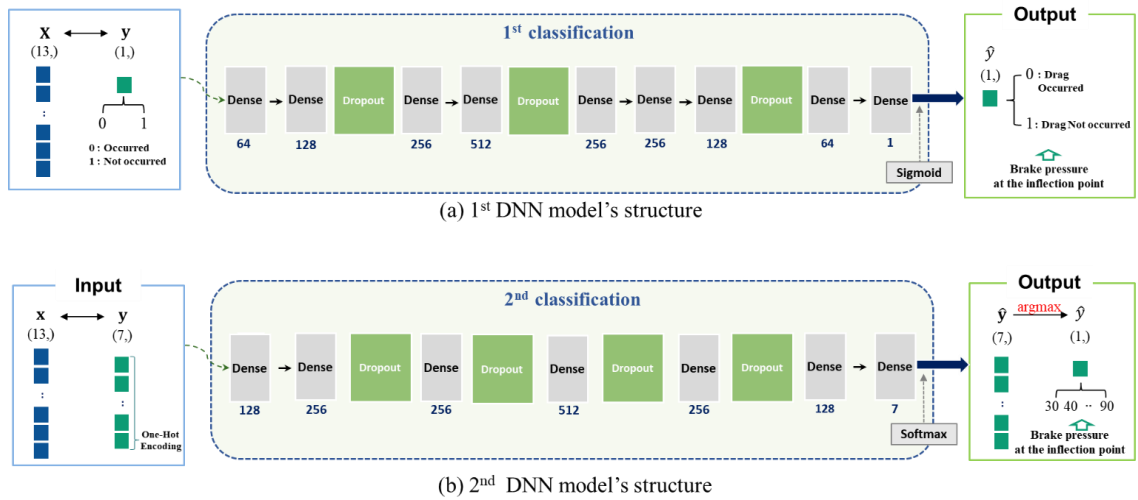

(a) 1st DNN model's structure

(b) 2nd DNN model's structure

Fig. 11. Architectures of the drag performance classification models.

Table 5. Drag performance classification modeling results.

| Methodology | Validation | Test |
| --- | --- | --- |
| First classification | 91.8 | 88.0 |
| Second classification | 68.2 | 65.8 |

# 6. SID (Stage 3a)

We developed the SID for APT only prior to developing the MID and verifying its feasibility. We compared the methodologies of finding the design value $x$ that could satisfy the target APT performance $y$ by using the APT prediction model. The three selected methodologies were SQP, backpropagation, and the proposed inverse design.

## 6.1. Inverse Design Model for APT

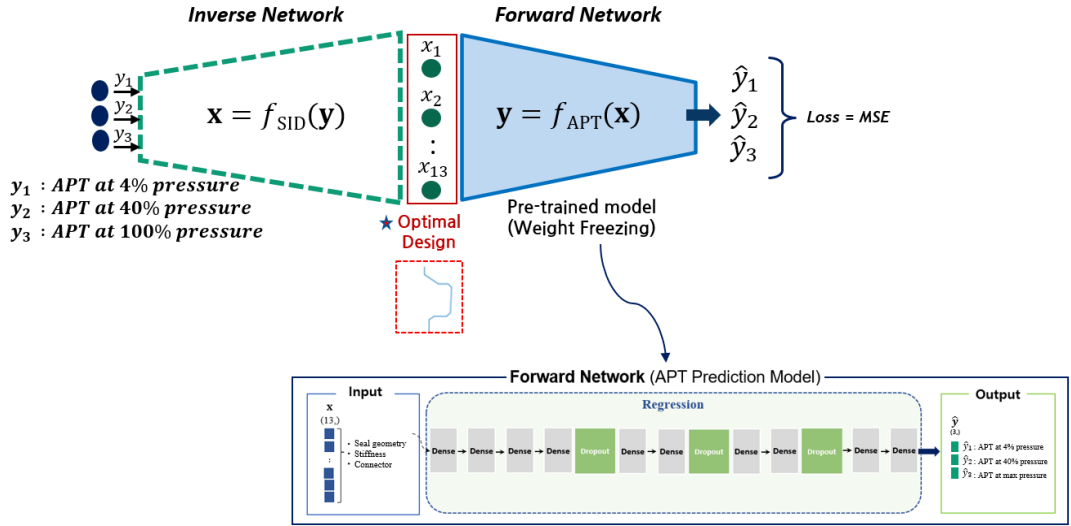

Fig. 12. SID architecture for APT.

Fig. 12 shows a DL architecture of SID for the APT. The structure used DNNs that combined the inverse network with the pretrained forward models. Here, we adopted the inverse design method of Liu et al. (2018), as explained in Section 2.2. The pretrained model for predicting APT ($y_1$, $y_2$ and $y_3$) with frozen weights was integrated into the inverse network. The inverse network was trained in a manner that the target APT ($y_1$, $y_2$, and $y_3$) would pass through the inverse network to generate 13 $x$'s and then enter the forward network again to find $x$ that could give the target $y$. For the inverse network, four fully connected layers with hidden layers were used. The hidden layers contained 256, 256, 128, and 64 neurons. The architecture of our inverse network was simpler than those of the forward networks. We learned empirically that the inverse network can be easily trained when the forward network performs excellently. An Adam optimizer with a learning rate of 0.0005 was used, and the batch size was set to 128. An early stop function was applied to prevent overfitting.

## 6.2. Baseline Optimization Models

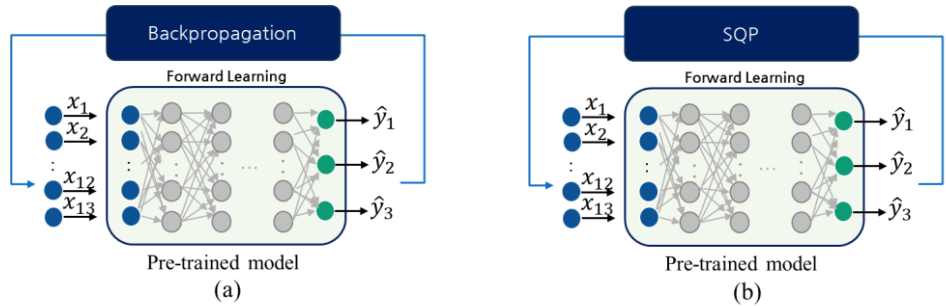

Fig. 13. Inverse design via iterative optimization using backpropagation and SQP.

We used SQP and backpropagation as the baselines for comparison with the inverse design (Fig. 13). The SQP methodology for iterative design is based on numerical first- and second-order derivatives, whereas the backpropagation methodology for design optimization is based on analytical first-order derivatives. However, iterative optimizers have a high computational cost and are highly dependent on the initial value, further suggesting that results may fall into the local optimum.

## 6.3. Design Optimization Results

The training dataset used in $\mathbf{x} = f_{SID}\,(\mathbf{y})$ consisted of 508 labeled pairs of data, with 146 (20%) and 73 (10%).

We used the same 73 test datasets to uniformly compare the optimization performance of the SID and baseline optimization models (i.e., backpropagation and SQP). After selecting the 73 APT performances of the test dataset, the objective function for the target performances was set to output 13 optimal variables. In this manner, the target performances could be achieved while performing optimization for non-constraint problems. MAE and RMSE were used to calculate the accuracy performance of the optimization. The performances are defined as follows:

$$MAE = \frac{1}{N}\sum_{i=1}^{N} |APT_{target_i} - APT_{optimized_i}|, \qquad (4)$$

$$RMSE = \frac{1}{N}\sum_{i=1}^{N} \sqrt{|APT_{target_i} - APT_{optimized_i}|^2}, \qquad (5)$$

where $i$ is the index of the target data among the 73 data, and $APT_{target_i}$ and $APT_{optimized_i}$ are the target and optimized performances in the $i$-th optimization. The calculations yielded 73 loss performances. Then, we calculated the average performance. The correlation between the target performance and optimized performance was also determined using $R^2$.

The values of the metrics used to evaluate the optimization performance of each model are summarized in Table 6. The SQP optimization model performed poorly (MAE = 0.073; RMSE of 0.107), whereas the SID model attained the best performance (MAE = 0.039; RMSE = 0.044). In addition, the optimized performance of the SID model showed an extremely high $R^2$ of 0.998.

The average values of computational cost of the SID and backpropagation models were 0.059 and 7.383 seconds (NVIDIA GeForce RTX 3090 D6X 24GB), respectively, whereas that of the SQP was 47.710 seconds (Intel I Core™ i5-8400 CPU at 2.80/2.81GHz). The least to highest computational cost is in the order of SQP, backpropagation, and SID. The SID model performed optimization in near-real time. The performance variables (accuracy and computational cost) of each optimization model are shown as a graph in Fig. 14. The inverse design model clearly outperforms the other two models in terms of accuracy and computational cost. In addition, some of the optimal design variables were verified in ABAQUS. The errors were approximately 0.0012 for MSE and 0.0346 for RMSE; these values are close to those of the target performances.

Table 6. Optimization modeling results.

| Methodology | SID | Backpropagation | SQP |
|---|---|---|---|
| MAE | **0.039** | 0.059 | 0.073 |
| RMSE | **0.044** | 0.067 | 0.107 |
| $R^2$ | **0.998** | 0.856 | 0.401 |
| Computational cost (seconds) | **0.059** | 7.383 | 48.710 |

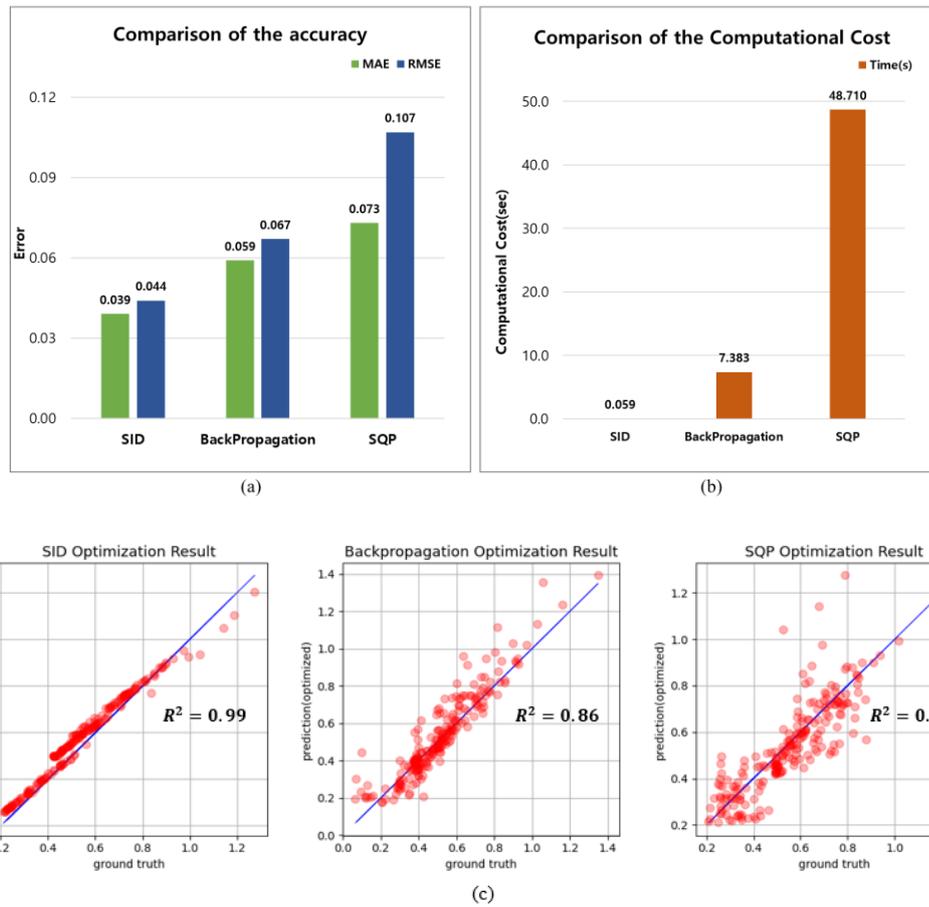

Fig. 14. Comparison of performance: (a) error, (b) computational cost, (c) $R^2$ for the SID, backpropagation, and SQP models.

Fig. 15 shows four examples of test data with detailed results from our SID model. The difference between the existing design and optimized design variables is large for the stiffness and connector variables, such as gp bush load limit ($x_{13}$), pad stiffness ($x_{11}$), and caliper stiffness ($x_{10}$), with values of 0.84, 0.54, and 0.48, respectively (Fig. 15(a)). This finding signifies the presence of a multiple local optima in the brake system design, further suggesting that different optimal designs can be used to meet the same target performance. Fig. 15(b) shows a comparison of the optimized seal geometry design and the existing seal geometry design. The result indicates that the optimized geometry differs from the existing geometry.

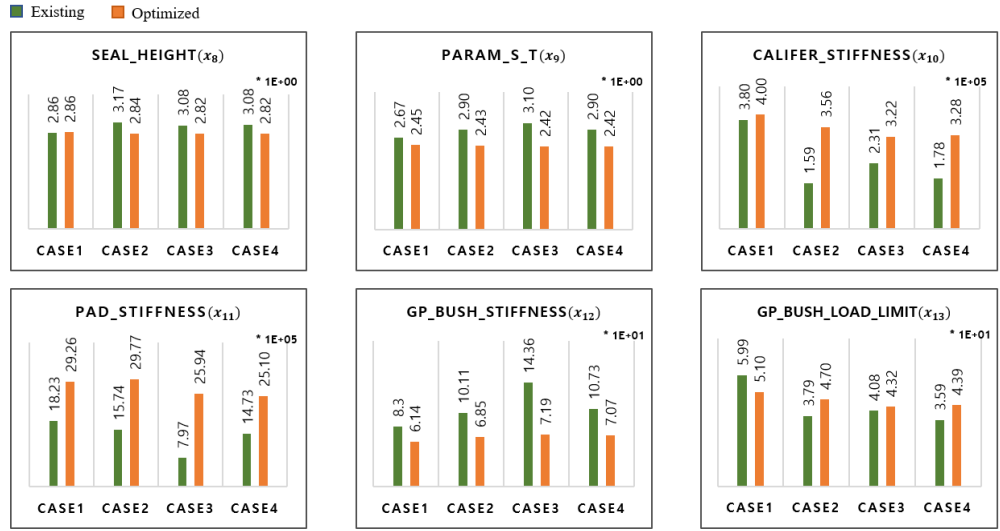

(a) Existing vs. Optimized Design variables I (Stiffness & Piston size) for SID

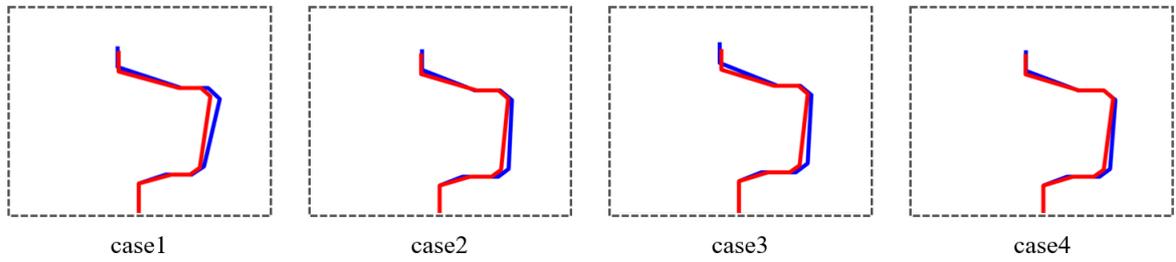

(b.1) generated seal geometry images overlaid on the existing shape

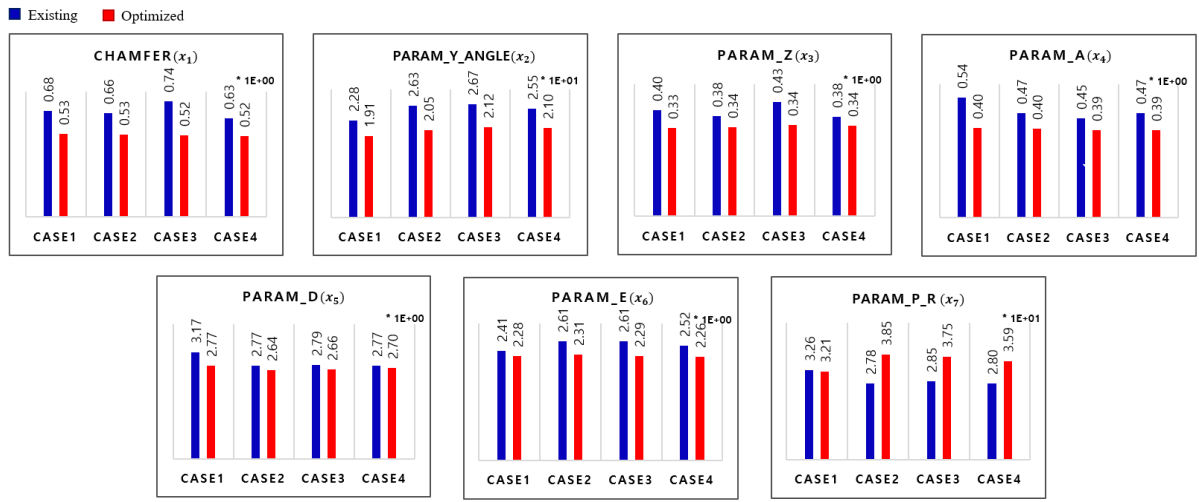

(b.2) Existing vs. Optimized Design variables II (Seal groove) for SID

Fig. 15. Detailed result of the test data samples for SID.

# 7. MID (Stage 3b)

7.1. MID Model

In complex system designs, such as the vehicle design, various design objectives or requirements from various disciplines need to be simultaneously satisfied, implying a tradeoff in the design goals of different teams (Kang et al., 2014). However, MDO has a high computational cost. Thus, the MID model is used for the real-time design optimization in this study to simultaneously satisfy the requirements of different disciplines.

In our study, two disciplines (APT and drag torque) are considered. A brake system has to meet the given target APT as much as possible, whereas drag torque should be avoided. In this case, APT can be used for the objective function, whereas the drag torque can be used as the equality constraint. (In Section 6, we considered a single discipline for satisfying the target APT, and the drag torque as a constraint was ignored.)

The MDO problem can be formulated as

$$\min_{\mathbf{x}} \quad \|APT_{target} - APT(\mathbf{x})\|^2 \\ subject\ to \quad Drag(\mathbf{x}) = 0 \quad (6)$$

The constrained optimization problem can be integrated into the neural network problem by converting the drag constraint into a regularization term, and then it can be added to the APT loss function. A multi-objective loss function can also be utilized, with the MSE loss used for the APT and the binary cross entropy loss for the drag constraint. Here, the APT has a continuous value, whereas the drag torque has a binary value. When a drag torque occurs, the value is 1; otherwise, 0.

The loss function is formulated as follows:

$$Loss = w_1 Loss_{APT} + w_2 Loss_{Drag}, \quad (7)$$

where $Loss_{APT} = \frac{1}{n}\sum_{i=1}^{n}(y_{APT,i} - \hat{y}_{APT,i})^2$,
$Loss_{Drag} = -\frac{1}{m}\sum_{i=1}^{m} y_{Drag,i}\log\hat{y}_{Drag,i} + (1 - y_{Drag,i})\log(1 - \hat{y}_{Drag,i})$,

where $y$ is the target value, and $\hat{y}$ is the predicted value. $w_1$ and $w_2$ are the weights for the multi-objective function; these weights can balance between the MSE loss and binary cross entropy loss. The empirical search process of the best weights is discussed in Section 7.2.

The proposed MID framework is shown in Fig. 16. The overall training process of MID can be described as follows.

First, we separately trained the APT and drag prediction models. By inputting design $\mathbf{x}$, the forward network could calculate the predicted values of $\hat{y}_1$, $\hat{y}_2$, and $\hat{y}_3$ passing through $y = f_{APT}(\mathbf{x})$ and the predicted value of $\hat{y}_4$ passing through $y = f_{Drag}(\mathbf{x})$.

Second, we added the two pretrained prediction models (forward network) in parallel after the inverse network. The input variables of the inverse network were the target APT and drag. In other words, in view of ensuring optimization via the inverse design, the output of the inverse network was used as the input of the two forward networks. $y_1$, $y_2$, and $y_3$ were selected as the targets for APT, and $y_4$ was selected as the target for drag in the inverse network, i.e., $x = f_{MID}(\mathbf{y})$. In this manner, the new design could be generated. Here, the target drag should always be 0.

Third, we froze the weights of the pretrained forward networks and trained only the inverse network such that the predicted results ($\hat{y}_1$, $\hat{y}_2$, $\hat{y}_3$, and $\hat{y}_4$) of the forward networks are closest to the input targets ($y_1$, $y_2$, $y_3$, and $y_4$). The same NN architecture of the SID was used for the inverse network of the MID. The learning rate was set to 0.00001, the batch size was set to 128. The Adam optimizer was used, and an early stop was adopted to prevent overfitting.

After the training, we used only the inverse network for inference. If target performances are given, then the inverse network can find the optimal design, and the targets can be satisfied in real time.

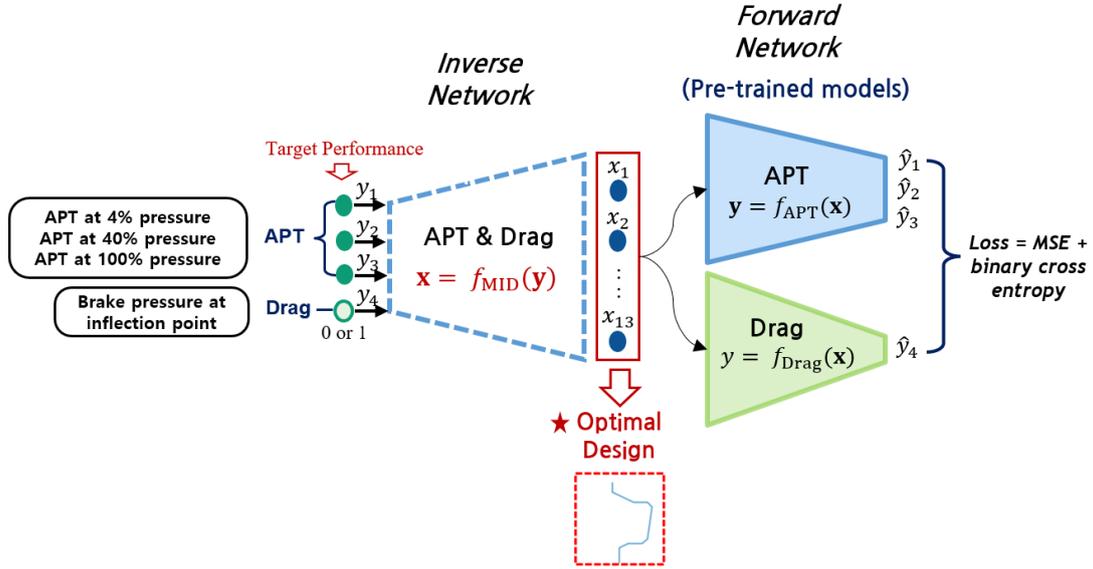

Fig. 16. MID architecture.

## 7.2. Design Optimization Results

The training dataset used in $\mathbf{x} = f_{MID}(\mathbf{y})$ consists of 508 labeled pairs of data, with 146 (20%) and 73 (10%). The SID model has an MAE of 0.038 and an RMSE of 0.044, as shown in Fig. 17(a). The MID model has an MAE of 0.041 and an RMSE of 0.050. The computational costs of the SID and MID models are 0.059 and 0.061 seconds (NVIDIA GeForce RTX 3090 D6X 24GB), respectively, as shown in Fig. 17(b). The computational cost is lower for the single-disciplinary inverse network that loads fewer forward models in contrast to the multidisciplinary inverse network, but both models can achieve the optimization results in near-real time. In addition, the optimized performances of the SID and MID model have a very high correlation with the target performance, with $R^2$ = 0.99 and 0.98, respectively. The value of the SID is slightly higher than that of the MID.

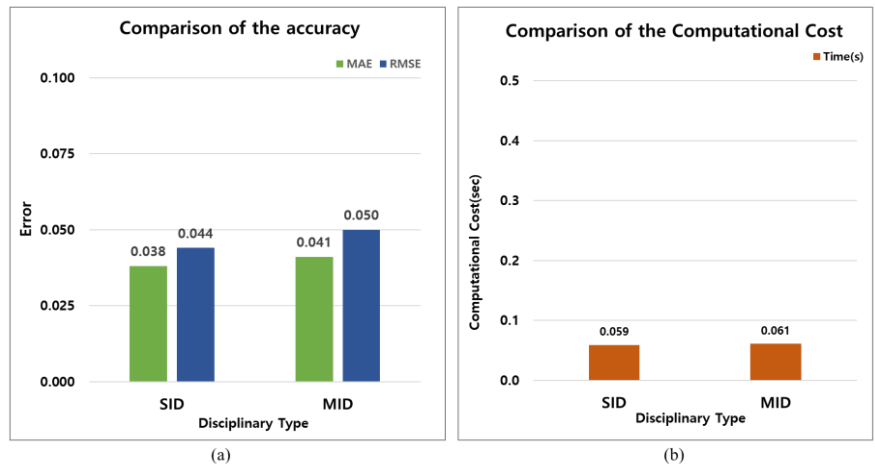

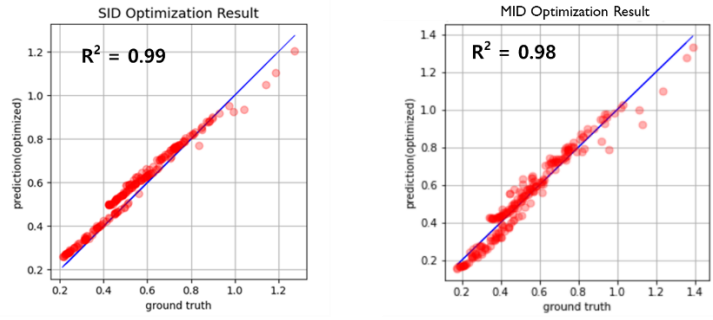

(c)

Fig. 17. Comparison of performance: (a) accuracy, (b) computational cost, and (c) $R^2$ of the optimization results of the SID and MID models.

Fig. 18(a) shows four examples of the test data with detailed results from our network. The difference between the existing design and optimized design values is large for the stiffness and connector variables, such as gp bush load limit $(x_{13})$, caliper stiffness $(x_{10})$, and pad stiffness $(x_{11})$, with values of 1.11, 0.79, and 0.50, respectively. This finding implies that the degree of design freedom for the input variables is as large as that of the SID. Thus, the local optimal designs can meet the target performances. Figs. 18(b.1) and 18(b.2) show the generated geometric designs with similar trends as that of the existing designs, but various values can be obtained.

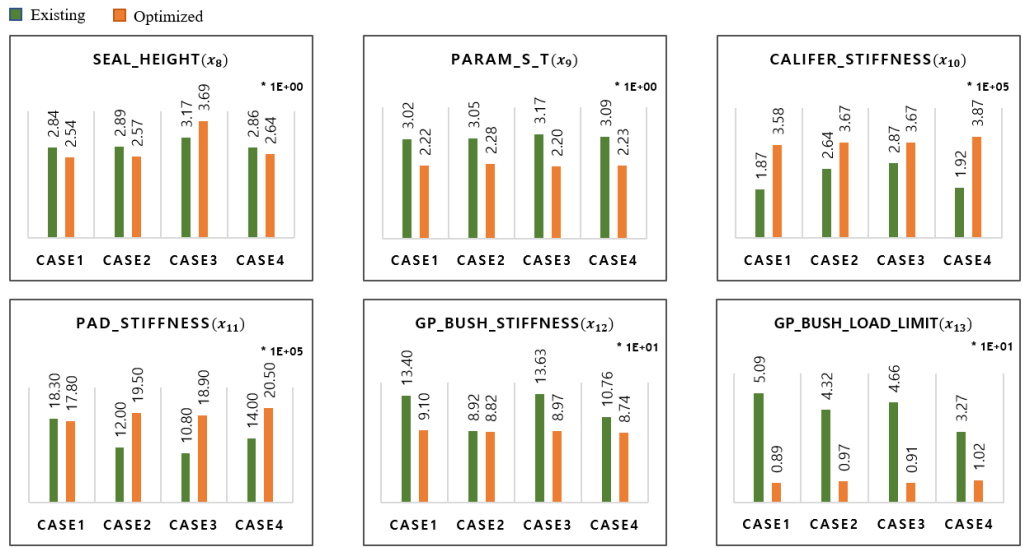

(a) Existing vs. Optimized Design variables I (Stiffness & Piston size) for MID

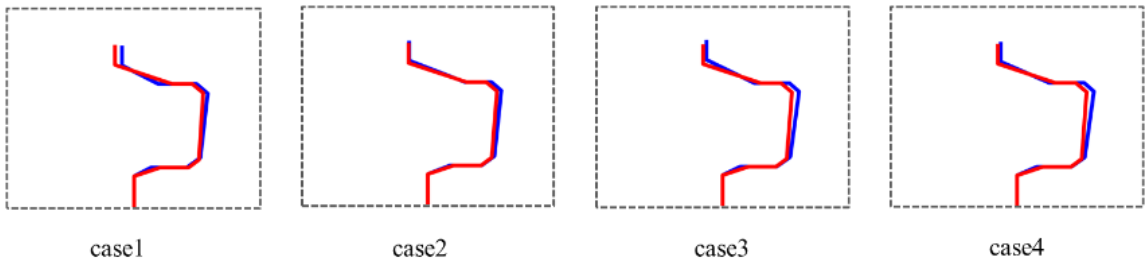

(b.1) generated seal geometry images overlaid on the existing shape

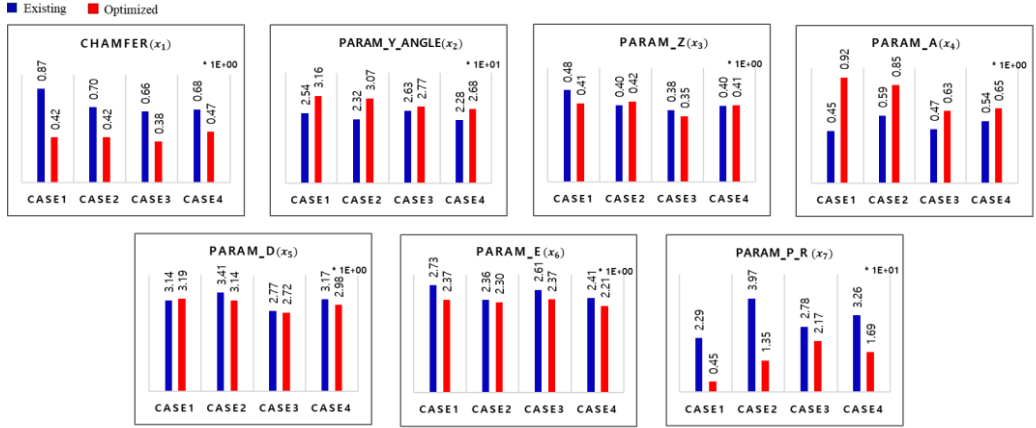

(b.2) Existing vs. Optimized Design variables II (Seal groove) for MID

Fig. 18. Detailed result of the test data samples for MID.

Table 7 shows the $Loss_{APT}$ and $Loss_{Drag}$ values pertaining to each weight of the loss function on the MID model. By gradually changing the weights of the loss in the objective function, we can empirically search the best weights. Here, before multiplying $w_1$ by $Loss_{APT}$, we additionally multiply 1e+03 by $Loss_{Drag}$ to adjust the scale. Although a smooth Pareto front is non-observable, the best two designs ⓒ and ⓔ are obtained. Designers can choose dot ⓒ or dot ⓔ depending on whether they want to focus more on objective A or objective B.

Table 7. RMSE and CEE values according to each weight.

| Dot | Weights | $Loss_{APT}$ | $Loss_{APT}$ |
|---|---|---|---|
| ⓐ | $w_1 = 0.6, \; w_2 = 0.4$ | 2.42e-03 | 4.27e-01 |
| ⓑ | $w_1 = 0.5, \; w_2 = 0.5$ | 2.30e-03 | 3.43e-01 |
| ⓒ | $\mathbf{w_1 = 0.4, \; w_2 = 0.6}$ | **2.31e-03** | **3.01e-01** |
| ⓓ | $w_1 = 0.3, \; w_2 = 0.7$ | 2.77e-03 | 4.55e-01 |
| ⓔ | $\mathbf{w_1 = 0.2, \; w_2 = 0.8}$ | **2.25e-03** | **3.09e-01** |
| ⓕ | $w_1 = 0.1, \; w_2 = 0.9$ | 3.02e-03 | 4.41e-01 |

## 8. Conclusion

This study proposes an MID framework for engineering systems to solve MDO problems in real time without the need to execute iterations. The scheme is demonstrated by considering a brake design problem that can simultaneously satisfy the target APT and drag constraint. The proposed process consists of three stages. In stage 1, the brake design and preprocess engineering simulation data are parametrized. In stage 2, the optimal model for predicting APT and drag performance are developed. In stage 3, the SID models and the MID model are developed on the basis of the pretrained models from stage 2, and the generated optimal design is verified by evaluating the target performance.

The contributions of this study and relevant insights can be summarized as follows.

First, to the best of our knowledge, this study is the first research in which the DL-based inverse design is applied to an MDO problem in the engineering field. We have shown that the MDO problem can be solved by transforming it into a multi-objective problem and learning it with an unconstrained loss function via DL. Previous inverse design studies have focused on nano/micro-level material research. Our study has demonstrated that system-level engineering design problems are suitable for DL-based inverse design approaches. DL-based inverse engineering is expected to become a popular design optimization approach that can be extended into many more engineering applications in the future.

Second, we have shown that the computational cost of the DL-based inverse design is remarkably lower than those of traditional optimization methods, such as SQP and backpropagation. Both SID and MID can achieve design optimization in real time (less than 0.1 second) after training the inverse network. However, the SID methodology has a better prediction on target performance than the SQP and backpropagation methodologies. As the concept design phase requires a consideration of various competing engineering performances, optimal concept designs must be immediately reviewed to attain a variety of target performances. Although this task is time consuming, the proposed methodology makes this possible in real time.

Third, we have shown that DL has real value when used for design optimization beyond surrogate modeling in the engineering field. Past engineering design studies have focused on surrogate modeling (metamodeling) by using DL. However, for general manufacturing with mass production, performing engineering simulations on demand is more efficient than training a DL-based surrogate model on large amounts of data. Design optimization requiring surrogate models and iterative optimization results entails a high computational cost. Thus, applying DL to design optimization is more necessary and effective than applying DL to surrogate modeling only.

The limitations of this study and future research plans can be summarized as follows. First, in supervised learning-based inverse design, only one design can be established for a target performance. We are currently conducting an inverse design research via unsupervised learning (Kim et al., 2021) to identify multiple optimal designs for satisfying various target performances. Second, the performance of inverse design is influenced by the dimensions and amount of data. We are currently conducting a study to compare the performance difference between the inverse design and existing optimization methodologies under various data scenarios (Jwa et al., 2021). In future research, the general guidelines for using DL-based inverse design will be provided, and through this initiative, the field of application of inverse engineering can be further expanded.

## Acknowledgements

This work was supported by Hyundai Mobis Company, VEng Company, and the National Research Foundation of Korea, with grants from the Korean Government (2017R1C1B2005266 and 2018R1A5A7025409).